# Evolutionary Approach to Test Generation for Functional BIST

Y.A.Skobtsov[1], D.E.Ivanov[2], V.Y.Skobtsov[2], R.Ubar[3], J.Raik[3]

[1]Donetsk National Technical University, Artema Str. 58, 83000 Donetsk, Ukraine,
skobtsov@kita.dongu.donetsk.ua
[2]Institute of Applied Mathematics and Mechanics of NAS of Ukraine, R.Luxemburg Str. 74, 83114 Donetsk, Ukraine
{ivanov | skobtsov}@iamm.ac.donetsk.ua
[3]Tallinn University of Technology, Raja 15, 12618 Tallinn, Estonia
{raiub | jaan}@pld.ttu.ee

**Abstract**

*In the paper, an evolutionary approach to test generation for functional BIST is considered. The aim of the proposed scheme is to minimize the test data volume by allowing the device's microprogram to test its logic, providing an observation structure to the system, and generating appropriate test data for the given architecture. Two methods of deriving a deterministic test set at functional level are suggested. The first method is based on the classical genetic algorithm with binary and arithmetic crossover and mutation operators. The second one uses genetic programming, where test is represented as a sequence of microoperations. In the latter case, we apply two-point crossover based on exchanging test subsequences and mutation implemented as random replacement of microoperations or operands. Experimental data of the program realization showing the efficiency of the proposed methods are presented.*

## 1. Introduction

According to ITRS roadmap [1], the number of test patterns to test a chip will explode to a size that will seriously affect test application times and thus, menufacturing costs. The roadmap shows that as the number of transistors per chip trends towards 180 million per square cm by 2012, the number of test vectors required will increase to 10 billion [2]. This amount of test data will be prohibitive, and the time to execute the test will reach around 10 seconds, even on a 1 GHz tester, pushing test times up by a factor of 50! Therefore, it is important to incorporate more design-for-testability (DFT) structures to the chip silicon itself in order to share the load between the tester and the hardware under test.

Functional, or arithmetic BIST approaches [3, 4] have been introduced to provide for an efficient solution to DFT for Systems-on-Chip. The main idea is to reuse the system infrastructure to test the core and additionally allow decompression of the external test set. However, the previous works on functional BIST have been mainly concentrated on decompression of a given set of test patterns [5, 6].

In current paper, we focus on automatic generation of tests for a system consisting of registers and combinational functional units. An evolutionary approach to test generation for functional BISTs is considered. Two methods of deriving a deterministic test set at functional level are suggested. The first method is based on the classical genetic algorithm with binary and arithmetic crossover and mutation operators. The second one uses genetic programming, where test is represented as a sequence of microoperations. In the latter case, we apply two-point crossover based on exchanging test subsequences and mutation implemented as random replacement of microoperations or operands.

The paper proposes a new functional BIST scheme with very low area overhead. The aim of the proposed scheme is to minimize the test data volume by allowing the device's microprogram to test its logic, providing a signature analyzer as an observation structure to the system. While system level, functional fault models are implemented in the experiments, fitness function based on logic level fault grading could be easily incorporated to the approach.

The paper is organized as follows. Section 2 explains the new functional BIST sheme used. Section 3 presents the genetic algorithm for functional test generation. Section 4 introduces genetic programming for generating tests in a functional BIST framework. Finally, experimental results and conclusions are given.

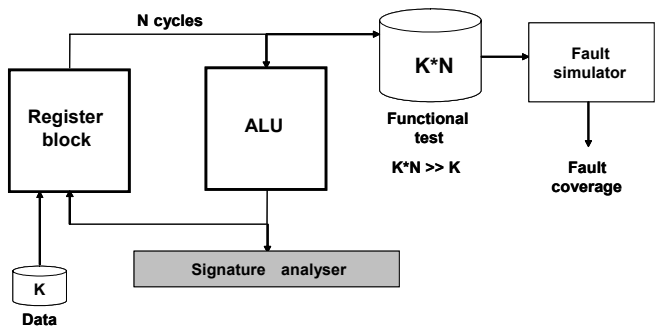

Fig. 1. A functional BIST scheme

## 2. The functional BIST scheme

Consider the datapath for the multiplication. It consists of a register block for storing the operands-multipliers, intermediate results of multiplication, the operation result, and the cycles counter. All necessary microoperations are carried out in the Arithmetic and Logic Unit (ALU), which has the role of circuit under test (CUT) (Fig.1). The ALU has data inputs and outputs connected via buses to the register block. The control signals from the control unit serve as additional inputs for ALU, and few status signals of the ALU serve as additional output signals connected to the control unit (not shown in Fig. 1).

During N cycles of the microprogram, ALU is exercised with N functional patterns, and the responses of ALU will be compressed in the signature analyzer which monitors the multiplication process. In the division process, we could use just K pairs of the operands A and B involved as the test for the ALU, and K quotients C = A·B as K responses to the test stimuli. However, in the FBIST scheme we will use all the K·N data words produced at the inputs of the ALU during the K·N cycles of the K multiplication operations as input stimuli to the ALU, and all the K·N data produced at the outputs of the ALU during the K·N cycles as the responses to stimuli. In such a way, we have got a multiplication effect of N times in the number of test patterns when moving the test access from the instruction level to the microinstruction level.

Denote by L the number of bits in the data (multiplication operands), and by l the number of bits on the inputs of ALU. The reduction in the test data volume through the compression of test data in the FBIST is equal to

$$R = \frac{Nl}{2L}.$$

For example (for the system used in the experiments), in the case of 32 bit words for the multiplier with 105 inputs and 120 cycles the reduction in the volume of test data is 120·105/64 = 197.

In this scheme the functional patterns produced directly at the inputs of ALU have similar role as pseudorandom test patterns in classical BIST schemes. Similarly to the pseudorandom test, the functional test patterns are not able to cover random-pattern-resistant faults, which limits the fault coverage that can be achieved with the pure functional BIST approach.

Let us consider functional test generation for the multiplier. Suppose that the operands $X$, $Y$ and the result $Z$ are integer numbers for simplicity (additional information about the bitwidth of the operands may be introduced). We shall apply two approaches to this problem:
1) a genetic algorithm (GA)-based method and
2) a genetic programming (GP)-based procedure.

## 3. GA for functional test generation

Genetic Algorithms (GA) are search algorithms based on the formalized principles of natural selection [7]. A subset of search space points, called a population, is chosen. Each potential solution of the problem: individual is presented by a chromosome with some gene structure. In the simplest case, an individual can be represented by binary encoded string. This makes GA attractive for solving the problem of logic circuits test generation, where the solution is presented as a set of binary patterns or sequences of binary patterns.

A fitness function is determined on the solution set and allows to estimate the closeness of each individual to the optimal solution: the ability to survive. The genetic search of solution consists of the simulation of such artificial population evolution. Creation of new individuals during the population evolution is based on the reproduction process simulation. In this case the individuals-solutions involved in reproduction process are called the parents, while the obtained individuals-solutions are called the offsprings. In each generation a set of individuals-solutions is constructed using the parts of individuals-parents and adding new parts with "good properties". Thus the GA effectively uses the information accumulated during evolution process. For solving any problem with genetic algorithm, first of all, we have to define:
1) the form of individual representation – encoding;
2) genetic operators – crossover and mutation;
3) fitness function.

Below we shall consider the definition of all GA components with the reference to the formulated problem. Here we use approach presented in [8].

1) A single test pattern consists of two integer numbers $X$ and $Y$ – the operands of multiplier. Therefore a decision is represented as two-component vector $(X,Y)$. On the other hand, we will use binary strings – codes of the numerical vectors, as decision representation.

2) Two types of genetic operators (crossover and mutation) are used – arithmetical and binary.

In accordance with arithmetical *crossover* for two parents $A = (X_a, Y_a)$ and $B = (X_b, Y_b)$, a new individual-offspring $\widetilde{A}$ is defined as follows

$$\widetilde{X} = (1-\alpha) \cdot X_a + \alpha \cdot X_b,$$
$$\widetilde{Y} = (1-\alpha) \cdot Y_a + \alpha \cdot Y_b,$$

where $\alpha \in [0;1]$.

Since we consider integer operands of multiplier and the described numerical crossover can generate floating point numbers, then we have to round off the generated components of the vector $\widetilde{A} = (\widetilde{X}, \widetilde{Y})$.

*Binary crossover* is executed according to the classical scheme represented below in Fig. 2

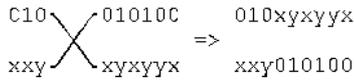

Figure 2: One-point binary crossover operator

Each type of crossover is applied with its own probability $P_b^c$ (binary) and $P_n^c$ (numerical): $P_b^c + P_n^c = 1$.

For *mutation operators* we use also two variations: arithmetical and *binary* mutation. Arithmetical mutation for functional testing is implemented as follows. New individual $A' = (X'_a, Y'_a)$, a *mutant*, is obtained from old individual $A = (X_a, Y_a)$ according to expressions

$$X'_a = X_a \pm \Delta \cdot X_a,$$
$$Y'_a = Y_a \pm \Delta \cdot Y_a$$

where $\Delta$ - is a small number. Obtained individual *A'* must be rounded off.

*Binary mutation* is implemented as a random bit inversion. Each type of mutation is applied with its own probability $P_b^m$ (binary) and $P_n^m$ (numerical): $P_b^c + P_n^c \ll 1$.

3) At the preliminary stage the test pattern quality is evaluated in the following way. The number of inverted bits in the multiplication result is estimated for each bit inversion in the current test pattern. The experiments have shown that a test pattern, where any bit inversion will lead to at least one bit inversion in the multiplication result, can always be found.

Our goal is now to generate test patterns such that the bit inversions of input operands produce maximum bit inversions in the multiplication result (it would be desirable to have all bit inversions). Thus we can define the matrix *P* of dimension $[2N \times M]$, where $p_{ij} = 1$, if *i*-th input bit inversion produces *j*-th output bit inversion. The matrix *P* is defined in the following way. First, all the matrix cells are zeros. Next, in the selected input pattern every bit is inverted. The matrix cells $p_{ij}$ are defined in accordance to the produced bit inversions in the output pattern. Thus, the fitness function is defined as

$$h(A) = \frac{\sum_{i=1}^{2N} \sum_{j=1}^{M} p_{ij}}{2NM}.$$

## 4. Genetic programming based functional test generation

GP-based approach to functional test generation is founded on the fact that multiplier can be represented not only at functional or structural levels, but at microprogrammed level too. In this case, multiplication procedure is described as microoperations sequence such as operand fetch, adding, shift etc. and test can be considered as a microoperations sequence. Under these conditions the methods of genetic programming can be effectively applied.

The modern genetic programming is one of the basic evolutionary computations paradigms and uses for individual representation the following three main forms:
1. Classic tree-like representation,
2. Linear graph representation,
3. Direct acyclic graph (DAG) representation [9].

Classical GP uses tree-like individual representation, which was good adopted for Lisp-programs. However, for example, for microoperations sequences or C-programs this form is not useful. For our purposes we offer to use second form – linear graph individual representation, because it is well applicable for representing individuals as microoperations sequence of variable length [10]. Note that all operations are executed on two operands − variables and constants, which form terminal symbols set. In this approach any operator is encoded with four-dimensional vector consisting of operation type and operands pointers.

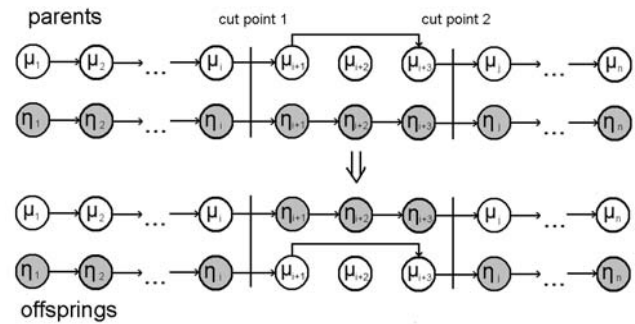

Fig.2. Microprograms crossover

Such kind of individual representation allows effective implementation of program recombination and their interpretation. In this case the tournament selection and two-point crossover are used. Here, two points are selected and parents are exchanged with the segment between selected points (Fig.2). Mutation is implemented as random replacement of operation type or operands values from given range.

## 5. Experimental results

The suggested algorithm was implemented in C++ language in the C++ Builder environment. The following main parameters of the GA-algorithm were empirically defined: values of crossover and mutation probabilities $P_c = 0.8$ and $P_m = 0.01$, coefficients $\alpha = 0.5$ and $\Delta = 0.5$ for functional crossover and mutation respectively, and the number of individuals in population was 100. Under these conditions the dependence of test fitness-function (fault coverage) from generations number was investigated. The experimental data in Fig. 6 show that fitness-function value is stabilized quickly enough. Thus, the boundary value of GA generations number is chosen equal to 40.

The results of the experiments to show the dependence of test fault coverage and test length on the operand bit capacity are shown in Fig.3. Average results of 10 experiments are cited.

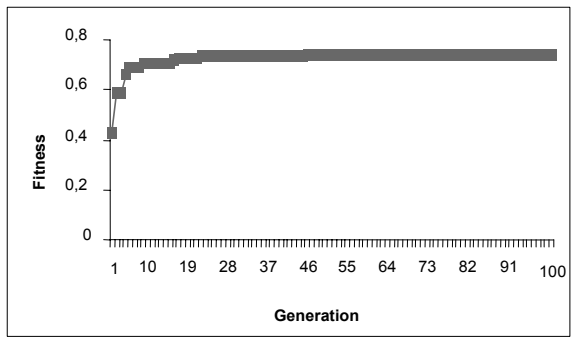

Fig.3. Growth of fitness-function value depending on number of generations

In order to evaluate the actual structural level fault coverage that can be achieved by the proposed functional BIST scheme, a case study with a microprogrammed datapath of a 32-bit floating point divisor was carried out. The combinational part of the datapath contains three 32-bit registers (dividend, divisor and quotient), a 5-bit counter, and a combinational ALU, which has 105 inputs, 71 outputs consisting of 513 gates and 2382 stuck-at faults, respectively.

Table 1 shows the experimental results obtained on that example. The experiments showed that seven operands were necessary to reach the maximum fault coverage of 89.8 per cent that could be achievable using the proposed BIST scheme. Higher fault coverage was not reachable because of a number of functionally untestable faults. A full-scan scheme would have covered those. This test of 7 operands takes 759 clock cycles to run. Since only the operands have to be saved to external or internal tester, the test compression rate achieved by the scheme is more than 100.

Table 1. Stuck-at fault coverage for the FBIST scheme

| k | operand1 | operand2 | result | $N_k$ | $N$ | FC, % |
|---|---|---|---|---|---|---|
| 1 | 0.7345 | 0.7659 | 0.9590 | 108 | 108 | 66.8 |
| 2 | 0.6943 | 0.7234 | 0.9598 | 105 | 213 | 76.7 |
| 3 | 0.4320 | 0.8569 | 0.5041 | 113 | 326 | 83.3 |
| 4 | 0.1964 | 0.2098 | 0.9361 | 108 | 434 | 85.5 |
| 5 | 0.4679 | 0.4987 | 0.9382 | 110 | 544 | 88.5 |
| 6 | 0.4567 | 0.4678 | 0.9763 | 104 | 648 | 88.9 |
| 7 | 0.9234 | 0.9546 | 0.9673 | 111 | 759 | 89.8 |
| .. | … | … | … | … | … | **89.8** |

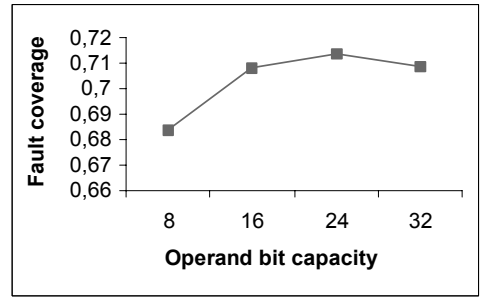

a)

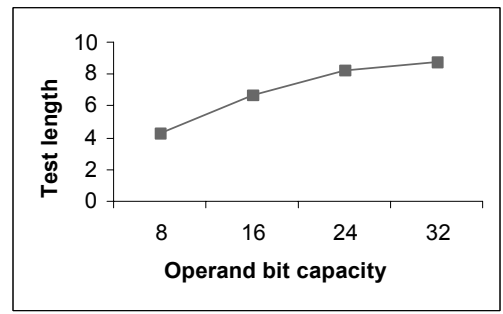

b)

Fig.4. Dependence of fault coverage and test length from operand bit capacity.

## 6. Conclusions and future work

The main goal of the proposed method is to generate as short as possible functional test to reduce the amount of input data for embedded functional BIST. The test generation for 32-bit multiplier with considered approach, shows that functional tests with 70% coverage relatively to primary input bits inversions were generated in a very low run time. We showed that the proposed functional BIST scheme allows to cover about 90 % of stuck at faults with a test data compression ratio of two orders of magnitude.

While experimental results were carried out on sequential multiplier and division algorithms, the proposed approach is open and applicable to any sequential arithmetic algorithm. The final fault coverage of the BIST can be increased to 100% with the test point insertion by path tracing method [11].

*Acknowledgements.* This work has been supported by EU V Framework projects REASON and EVIKINGS, as well as by Technology Development Center ELIKO and Estonian Science Foundation grants 5649, 5910 and 5637.

## References

[1] The International Technology Roadmap for Semiconductors, 2001 edition. International Sematech, Austin, Texas, 2001.

[2] R. Leckie. The Test Technology Roadmap. *Semiconductor Fabtech* – 8th edition, pp. 309-314.

[3] J. Rajski, J. Tyszer, *Arithmetic Built-in Self-test for Embedded Systems*, Pearson Professional Education, Nov. 1997, p. 256.

[4] R. Dorsch, H.-J. Wunderlich. Accumulator based deterministic BIST, In Proc. Int. Test Conf., 18-23 Oct. 1998, pp. 412 – 421.

[5] R. Dorsch, H.-J. Wunderlich. Reusing scan chains for test pattern decompression, In Proc. Int. Test Conf., May 29 - Jun. 1, 2001, pp. 124 – 132.

[6] O. Novak, J. Nosek. Test pattern decompression using a scan chain, Proc. IEEE International Symposium on Defect and Fault Tolerance in VLSI Systems, 24-26 Oct. 2001, pp. 110 – 115.

[7] D.E. Goldberg, Genetic Algorithms in Search, Optimization & Machine Learning. Addison-Wesley Publishing Company, Inc., 1989.

[8] Y.A. Skobtsov, D.E. Ivanov, V.Y.Skobtsov, R. Ubar "Evolutionary approach to the functional test generation for digital circuits". In Proc. Of the $9^{th}$ Biennial Baltic Electronics Conference, 2004, pp. 229-232.

[9] W. Banzhaf, P. Nordin, R. E. Keller, and F. D. Francone. "Genetic Programming: An Introduction." Morgan Kaufmann, Inc., San Francisco, USA, 1998.

[10] W. Kantschik, W. Banzhaf, "Linear-Graph GP -- A new GP Structure", EuroGP2002: 4 tn European Conference on Genetic Programming, 2002, pp. 83-92.

[11] N.A. Touba, E.J. McCluskey, "Test point insertion based on path tracing". In Proc. of IEEE VLSI Test Symposium, 1996, pp.2-8.